\documentclass{article}

\usepackage[final]{corl_2022} 

\usepackage{amsmath}
\usepackage{amssymb}
\usepackage{mathtools}
\usepackage{amsfonts}
\usepackage{algorithm}
\usepackage{algpseudocode}
\usepackage{graphicx}
\usepackage{bbding}
\usepackage{pifont}
\usepackage{wasysym}
\usepackage{mathtools}
\usepackage{wrapfig} 
\usepackage{bm}
\usepackage[colorinlistoftodos]{todonotes}
\usepackage{booktabs}
\usepackage{subcaption}
\usepackage{booktabs,caption,blindtext}
\usepackage{tocbibind}
\usepackage[toc,page]{appendix}
\usepackage{etoc}

\newcommand{\eg}{e.g., }
\renewcommand{\vec}[1]{\bm{#1}}

\newcommand {\rb}[1]{{\color{black}\textbf{}#1}\normalfont}

\usepackage{tikz}

\makeatletter
\renewcommand{\fnum@figure}{\small Fig.~\thefigure}

\makeatother

\newcommand{\accro}[0]{VIRDO++~}
\newcommand{\cntemb}[0]{ \textcolor{black}{contact latent vector}}
\title{VIRDO++: Real-World, Visuo-Tactile Dynamics and Perception of Deformable Objects}

\author{
Youngsun Wi$^1$\hspace{15pt} Andy Zeng$^2$\hspace{15pt} Pete Florence$^2$\hspace{15pt} Nima Fazeli$^1$ \\
$^1$Robotics Department, University of Michigan \hspace{15pt}$^2$Robotics at Google \\
\texttt{\{yswi, nfz\}@umich.edu} \hspace{15pt} \texttt{\{andyzeng, peteflorence\}@google.coms}\\
\url{https://www.mmintlab.com/virdopp}
}

\begin{document}
\maketitle

\begin{abstract}
    Deformable objects manipulation can benefit from representations that seamlessly integrate vision and touch while handling occlusions. 
    In this work, we present a novel approach for, and real-world demonstration of, multimodal visuo-tactile state-estimation and dynamics prediction for deformable objects. Our approach, VIRDO++, builds on recent progress in multimodal neural implicit representations for deformable object state-estimation \cite{young2022virdo} via a new formulation for deformation dynamics and a complementary state-estimation algorithm that (i) maintains a belief \rb{distribution of deformation within a trajectory}, and (ii) enables practical real-world application by removing the need for \rb{contact patches.} 
    In the context of two real-world robotic tasks, we show: (i) high-fidelity cross-modal state-estimation and prediction of deformable objects from partial visuo-tactile feedback, 
    and (ii) generalization to unseen objects and contact formations. 
\end{abstract}


\keywords{\small Deformable Object Manipulation, Multimodal Representation Learning} 


\begin{figure}[thpb]
  \centering
    \includegraphics[width=1\textwidth]{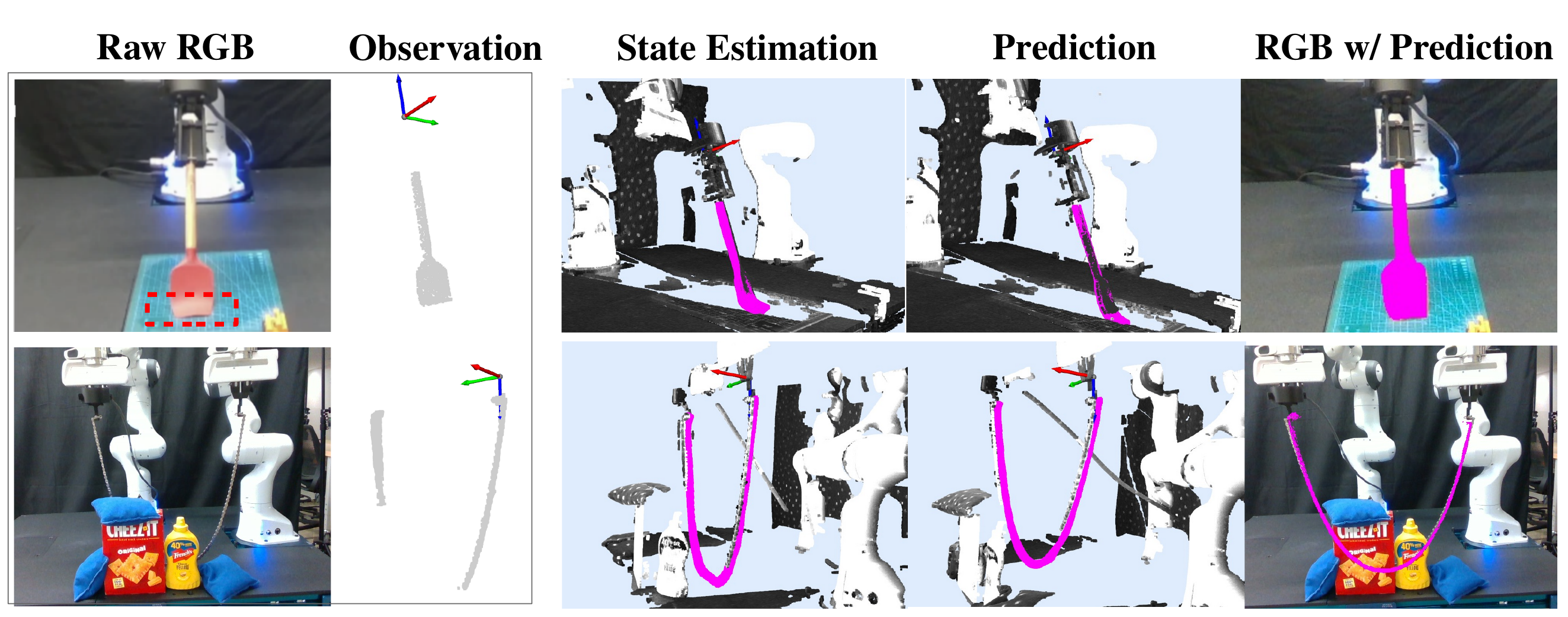}
      
      \caption{\small
      Given partial point cloud observations (left), and multimodal sensor measurements (end effector wrist reaction wrench and pose), \accro can accurately predict the 3D geometry (\textcolor{magenta}{magenta}) and reaction forces of deformable objects such as spatulas (top) and bike chains (bottom) conditioned on the next robot action under severe occlusions -- which can be artificial (\textcolor{red}{red dotted box}) or natural due to obstacles (\eg from other objects).
      }
    \vspace{-15pt}

  \label{fig:teaser}
\end{figure}

\section{Introduction}
\vspace{-5pt}

Deformable objects are ubiquitous in our everyday lives -- the clothes we wear, the food we eat, and many of the tools we use are just a few examples. As such, helpful robots of the future may benefit from the ability to master dexterous manipulation of deformable objects. 
At the heart of this mastery is the interplay between object geometry and force transmission, perceived by vision and touch. Deformable object manipulation is difficult due to the complexity of this interplay (e.g., infinite dimensional state-spaces and nonlinearity) and the ensuing challenges in perception (e.g., partial observability and occlusion) and controls \cite{yin2021modeling,sanchez2018robotic,arriola2020modeling,hou2019review}. Multimodal visuo-tactile representations can help address many of these challenges by exploiting mutual information and complementary cues. 

Existing deformable object manipulation approaches typically use one modality (mostly vision) 
and rely on finite element/particle-based techniques \cite{essa1992unified,ficuciello2018fem,frank2014learning, muller2004interactive,sengupta2019tracking,howard2000intelligent,nurnberger1998problem,tonnesen2000dynamically} or leverage deep learning for visual affordance/latent dynamics learning \cite{thach2021learning,thach2021deformernet,weng2022fabricflownet,lin2022diffskill,seita2018deep,hoque2021visuospatial}. The former methods typically rely on privileged knowledge (e.g., occluded or unknown boundary conditions) and stop at system identification, limiting their ability to refine the underlying physics model by learning from data.
Latter methods lack i) tactile feedback which is far more informative of contact than vision, and ii) structured representations which can limit generalization and increase sample-complexity.


Here, we build on recent work \cite{young2022virdo} that provides
a framework for visuo-tactile state estimation using multimodal implicit neural representations. This approach offers a number of advantages including multimodal sensory fusion, direct integration with raw sensory feedback, and computational speed. However, \cite{young2022virdo} does not address 1) deformation predictions (dynamics) and 2) is not directly applicable to real-world robotics settings. The first limitation is due to the lack of a dynamics module and an emphasis on static scenes. The second limitation is due to the need for the contact patch which are typically impractical to measure in real-world robotic settings. 

The primary contribution of this work is (i) a novel approach to, and (ii) real-world demonstration of, performing multi-modal visuo-tactile state estimation and dynamics prediction for deformable objects. Fig.~\ref{fig:teaser} shows how our proposed approach uses partial views and tactile feedback for deformation prediction and state-estimation in real-world applications. Our contributions are:

\textbf{1. Representation of deformable objects \textit{dynamics} conditioned on actions:} Multimodal neural implicit representation with action-conditioned dynamics to predict future deformations. To demonstrate the utility of the dynamics model, we introduce a state estimation algorithm based on particle filtering to maintain a belief distribution over object deformations.

\textbf{2. Real-world demonstrations on two challenging tasks:} Scraping with a spatula and chain manipulation. We demonstrate both 3D geometry estimation in heavy occlusion and action-conditioned deformation/reaction wrench prediction. This is enabled by introducing a novel \cntemb~ that eliminates the need for explicit contact patch information used in \cite{young2022virdo}. We use these tasks to demonstrate \textbf{generalization to unseen objects and novel environments}.

\vspace{-10pt}
\section{Related Work} \label{sec:related}
\vspace{-10pt}

\textbf{Deformable object modeling.} Recent studies have attempted to address challenges in system identification and high computation costs in conventional continuum mechanics \cite{essa1992unified,ficuciello2018fem,frank2014learning, muller2004interactive,sengupta2019tracking,howard2000intelligent,nurnberger1998problem,tonnesen2000dynamically}. To ease the burden of system identification, prior studies infer the physics models' parameters based on high-fidelity physics engines \cite{sundaresan2022diffcloud,Antonova2022Bayesian,narang2022factory} and simple force-deformation relationship (e.g. Hooke's law) \cite{ petit2018capturing}. The inherent limitations of these approaches are that i) performance is confined to the underlying physics model, often with strong assumptions on the objects (uniform density/elasticity) and the force-deformation relationship (linearity), and ii) access to privileged information such as occluded or unknown boundary conditions. More recent approaches propose computationally efficient elastic object modelings with minimum system identification using, for example, a potential energy propagation \cite{zhang2019neural} and geometric motion estimates \cite{wang2021tracking}. However, these methods suffer from balancing between model approximation and computation cost as well as maintaining seamless integration with robotic sensing modalities (RGB-D cameras and F/T sensors). In this study, we adopt a data-driven approach using neural networks \cite{young2022virdo, Palafox2021ICCV,ma2021learning}. Learning directly from observation, our approach does not require object parameters (Young's modulus, Poisson's ratio) or a strong assumption of object composition. Even so, it can capture complex non-linear elastic behaviors and exhibit fast computation times due to the architecture of the neural networks.

\textbf{3D geometric representation of deformable objects.} Most recent studies have adopted discrete geometry representation for deformable objects with finite resolution such as meshes \cite{ficuciello2018fem,sengupta2019tracking,Santesteban2021CVPR} and key points \cite{thach2021deformernet, Antonova2022Bayesian}, sometimes accompanied with graph neural networks \cite{shi2022robocraft,ma2021learning}. \rb{These approaches have been used largely for 1D and 2D objects (rope, cloth, top view of cylindrical objects). Among them, structured representations (e.g., Mesh or GNN) are advantageous at tracking connectivity, mainly used for deformable objects with large self-occlusions (e.g. folded cloth or knot)  \cite{huang2022mesh, chi2021garmentnets}}. On the other hand, we address general 3D geometry representations which are not confined to specific types of objects and can work with large occlusions (e.g., wok-scraping). Our work builds on dense implicit geometric representation which include signed distances fields~\cite{young2022virdo,Palafox2021ICCV,ortiz2022isdf}, occupancy~\cite{Niemeyer2019ICCV}, volume density~\cite{park2021nerfies}, and dense descriptors~\cite{sundaresan2020learning}. These approaches have several advantages over their discrete counterparts in dealing with large occlusion, high resolution surface representation \cite{sitzmann2020implicit}, parameterization of 3D geometries \cite{park2019deepsdf, DIF}, and are useful for downstream tasks (e.g., contact location detection similar to Sec.~\ref{sec: bike chain demo}). We extend these representations to real-world multimodal deformable object dynamics and perception by integrating touch and robot actions.



\vspace{-5pt}
\section{Methodology}
\vspace{-5pt}

\begin{figure}
    \centering
    \includegraphics[width=0.9\textwidth]{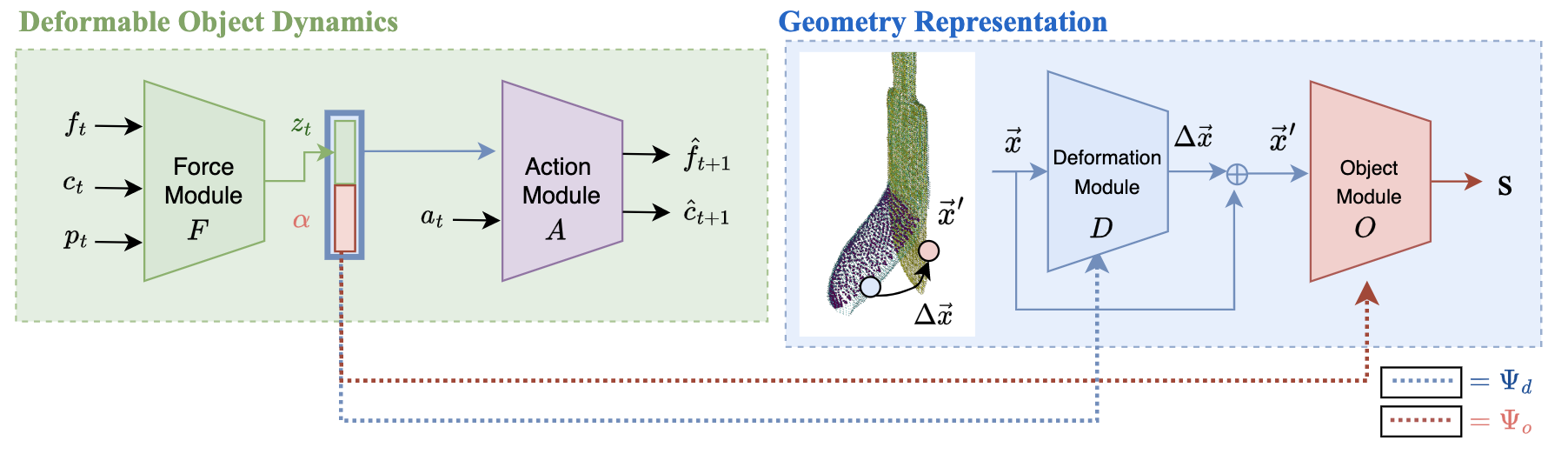}
    \vspace{-10pt}
    \caption{\small\textbf{Overview.} \rb{\accro is composed of an implicit signed-distance field representation of geometry (left), informed by a deformation dynamics model (right), where dotted lines indicate hyper networks that decode embeddings into network weights. \label{Fig: network diagram}}}
    \vspace{-15pt}
\end{figure}
Our approach, shown schematically in Fig.~\ref{Fig: network diagram}, is composed of a latent deformation dynamics model and an implicit dense geometric representation. Using Hidden Markov Models as an analogy, the former component plays the role of the hidden state transition, and the latter is the observation model. We expand on the latent deformation dynamics in Sec.~\ref{sec: def-dynamics} and discuss the implicit geometric representation in Sec.~\ref{sec:geometry represention}. 

\vspace{-5pt}
\subsection{Deformable Object Dynamics \label{sec: def-dynamics}}
\vspace{-5pt}

\accro represents deformable object states using latent object ($\vec{\alpha}$) and force ($\vec{z_t}$) codes. The object code condenses the undeformed shape information into a feature representation that can extrapolate to new unseen object variants. The force code is generated by the \textcolor{LimeGreen}{Force Module ($\mathbf{F}$)} which encodes boundary conditions as $\vec{z}_t = \mathbf{F}(\vec{f}_t, \vec{c}_t, \vec{p}_t)$ where $\vec{f}_t \in \mathbb{R}^{6} $ is the reaction wrench at the wrist, $\vec{p}_t \in \mathbb{R}^{6}$ is robot end-effector pose, and $\vec{c}_t \in \mathbb{R}^{lc}$ is the \cntemb \rb{, where $lc$ is the dimension of the \cntemb.} The \cntemb~ $\vec{c}_t$ is a learnable latent vector that can change over time whose primary role is to store contact information. However, it can also implicitly integrate information about the object's physical properties (e.g., stiffness) and historical information about the object's states and deformations. This embedding generalizes the original VIRDO \cite{young2022virdo} in terms of both (i) no longer needing known contact patch locations, unlocking the capability to run VIRDO in the real-world, and (ii) representing additional useful information beyond just the contact location as was the case in VIRDO. In more detail, VIRDO \cite{young2022virdo} used ground truth contact patches encoded by a PointNet encoder \cite{qi2017pointnet} directly which required access to this privileged information. Here, the contact patch and encoder are replaced by the learnable \cntemb.



The \textcolor{violet}{Action Module ($\mathbf{A}$)} predicts the next step's boundary conditions given the current (latent) state and robot action: $\mathbf{A}( \vec{\alpha}, \vec{z_t}, \vec{a_t}) =  \hat{\vec{f}}_{t+1},  \hat{\vec{c}}_{t+1}$ where  $\hat{\vec{f}}_{t+1}\in \mathbb{R}^{6}$ is the predicted reaction wrench at the wrist, $\vec{\hat{c}}_{t+1}$ is the predicted \cntemb, and $\vec{a}_t \in \mathbb{R}^{6}$ is the robot action (Cartesian displacement). \rb{This predictive capability enables state-estimation and downstream tasks such as
contact location detection crucial for deformable object manipulation. Though we did not include in this paper, the predictive ability of \accro also enables planning for deformable object manipulation.} 
VIRDO lacked this functionality and could only infer the object geometry statically.

\vspace{-5pt}
\subsection{Geometry Representation \label{sec:geometry represention}}
\vspace{-5pt}
\accro decouples the 3D geometric representation of deformable objects into an undeformed (nominal) shape signed-distance field (SDF) and a set of deformation fields, similar to \cite{young2022virdo, DIF}. The deformation field is a 3D vector field that, when summed with the deformed object SDF, results in the nominal shape SDF. More precisely, any query point $\vec{x} \in \mathbb{R}^{3}$ in the robot wrist frame belonging to the deformed shape SDF can be mapped back to the corresponding point $\vec{x}'$ of the nominal shape SDF by applying the point-wise deformation field $\Delta \vec{x}$ as $ \vec{x} + \Delta \vec{x} = \vec{x}' $. In more detail:




\textbf{Nominal Shape Representation: \label{sec:nominal shape representation}} The \textcolor{red}{Object Module $\mathbf{O}(\vec{x}')$} is the parametric representation of the nominal shape SDF -- the object shape with no external contacts corresponding to $\Delta\vec{x} = 0$. Here, we use a feedforward neural network to parameterize the SDF \cite{park2019deepsdf} and write $\mathbf{O}(\vec{x}'|\mathbf\Psi_o(\vec{\alpha})) = s$ where $s$ is the signed-distance at query point $\vec{x}'$ and $\mathbf\Psi_o(\vec{\alpha})$ is a hyper-network that predicts the weights of $\mathbf{O}$ conditioned on the object code $\vec{\alpha}$. The object code and hyper-network weights are learned end-to-end in an auto-decoder approach \cite{park2019deepsdf}. We simplify our notation to $\mathbf{O}(\vec{x}')$ for the remainder.


\textbf{Deformation Field Representation: \label{sec: Deformed Shape Representation}} The generalized geometric representation of a deformable object is given by $\text{SDF}(\vec x) = \mathbf O_{\mathbf{\Phi}_o} \big( \vec x + \mathbf{D}( \vec{x}| \Psi_d( \vec{z}_t, \vec{\alpha}_t)) \big)=s$. Here, the \textcolor{Aquamarine}{Deformation Module} $\mathbf{D}( \vec{x}| \Psi_d( \vec{z}_t, \vec{\alpha}_t)) = \Delta \vec{x}$ produces an object's deformation field given force and object code pairs. Similar to the Object Module, Deformation Module's weights are given by the hyper-network $\Psi_d$. 

\vspace{-5pt}
\subsection{Training and Loss Formulation \label{sec: loss formation}}
\vspace{-5pt}
The training dataset for \accro is a set of trajectories $\mathcal{S}=\{ \mathcal{T}_1,\mathcal{T}_2, ..., \mathcal{T}_N \}$, where each trajectory $\mathcal{T}_j = \{ (\vec{p}_0, \vec{f}_0, \vec{P}_0, \vec{a}_0), (\vec{p}_1, \vec{f}_1, \vec{P}_1, \vec{a}_1), ... , (\vec{p}_k, \vec{f}_k, \vec{P}_k, \vec{a}_k) \}$ has $k$ sequential observation and action tuples. Here $\vec{P}_j$ is a partial point cloud. Our first step is to pretrain a nominal shape representation (i.e., hyper-network $\mathbf{\Psi_o}$ and object code $\vec \alpha$). The loss for training $m$ nominal shapes is $\mathcal{L}_{nominal} =  \mathcal{L}_{sdf} + \lambda_1  \mathcal{L}_{latent} + \lambda_2 \mathcal{L}_{hyper} $ as in \cite{young2022virdo}, where $\mathcal{L}_{latent} =\sum_{i=1}^m \| \vec{\alpha}^i \|_2$, $\mathcal{L}_{hyper} =\sum_{i=1}^m \| \mathbf\Psi_o(\vec{\alpha}^i) \|_2$, and:
\begin{equation}
\mathcal{L}_{sdf} =\sum_{i=1}^m \Big(
\sum_{\bar{\vec{x}} \in \mathbf{\Omega}} |clamp(\mathbf{O}^i(\vec{x}), \delta)- clamp(s^*,\delta)| \nonumber + \lambda\sum_{\bar{\vec{x}} \in \mathbf{\Omega}_{0}}(1-\langle \nabla \mathbf{O}^i(\vec{x}), \vec{n}^*\rangle)\Big).
\label{eq: pretrain loss}
\end{equation} 
$\vec{n}^*$ is the ground truth normal vector, $\delta$ is a parameter that clips SDF predictions, and $s^*$ is the ground truth signed distance. After the pretraining, we train the rest of the modules using the three losses $\mathcal{L}^{geo}, \mathcal{L}^{pred}$ and $\mathcal{L}^{reg}$ summed over a fixed horizon $(w)$ while freezing $\Psi_o$ and the object code $\alpha$: 
\begin{equation}
\mathcal{L}_{tot} = \sum_{t=t_0}^{t_0+w}(\mathcal{L}^{geo}_{t}+ \lambda_3 \mathcal{L}^{pred}_{t}+\lambda_4 \mathcal{L}^{reg}_{t}) \label{eq: L_dynamics}
\end{equation}

The geometry representation loss is defined as:
\begin{equation}
\begin{aligned}
 \mathcal{L}^{geo} =
\lambda_{5}\underbrace{\sum_{\vec x \in \mathbf \Omega}\| \mathbf D_{\mathbf\Psi_d}(\vec x) \|_2}_{\text{minimum correction}} &+
\lambda_{6}\underbrace{\text{CD}(\vec P + \mathbf D_{\mathbf\Psi_d}(\vec P), \bar{\vec{P}}^*)}_{\text{correspondence}} +
\lambda_{7}\underbrace{\sum_{\vec{x} \in \mathbf{\Omega}_{0}}(1-\langle \nabla_{\vec{x}} \mathbf{O}_{\Psi_o}(\vec x'), \vec{n}^*\rangle}_{\text{normal aligning}} \\
&+\lambda_{8}\underbrace{\sum_{\vec{x} \in \mathbf{\Omega}}| clip( \mathbf O_{\mathbf\Psi_o} \big( \vec x + \mathbf D_{\mathbf\Psi_d}(\vec x), \delta \big) - clip(s^*, \delta )|}_{\text{signed distance regression}}.
\end{aligned}
\label{eq: L_geo}
\end{equation}
where $\vec x' = \vec x + \mathbf{D}_{\Psi_d}(\vec x)$, $\Omega$ is the 3D querying space,  $\Omega_0 \subset \Omega$ is the on-surface region,  $\mathbf{P}:=\{\vec{p}| \vec{p} \in \Omega_0\}$ is an unordered set of the on-surface points, and  $\bar{\vec{P}}^*$ is the ground truth nominal shape point cloud. Next, we add a reaction wrench prediction loss as $ \mathcal{L}_{pred} = \| \vec{f}_{t+1} - \hat{\vec{f}}_{t+1} 	\| $. The prediction loss for $\hat{\vec{c}}_{t+1}$ is implicitly handled by Eq.~\ref{eq: L_geo} because the \cntemb~ prediction is recursively used as next step's input during training. In contrast, we do replace $\hat{\vec{f}}_{t+1}$ with real measurement in Eq.~\ref{eq: L_geo}. Finally, regularization losses $\mathcal{L}^{reg}_{t}= \lambda_9 \| \vec{z}_t \|_2 +   \lambda_{10}\| \vec{c}_t \|_2 +  \lambda_{11}\| \mathbf\Psi_d(\vec{\alpha}, \vec{z}_t) \|_2$, which regularize the force codes, \cntemb, and the weights of $\mathbf{D}$ respectively. 

\vspace{-5pt}
\subsection{Inference and State-estimation Algorithm  \label{sec: state estimation}}
\vspace{-5pt}

To maintain a belief distribution over the object deformation, we develop a particle filter-based inference algorithm, Alg.~\ref{Alg: particle filter}. Each particle represents a \cntemb~ which can be used to reconstruct the full 3D object geometry. The $\texttt{Refine}$ step updates the particle set by performing gradient descent on $\min_{\mathbf{\hat C_t}} \mathcal{L}^{geo}(\vec{P}_t,\mathbf {\hat{C}}_t, \vec{f}_t, \vec{p}_t)$ with the visible object point cloud. The particles are then weighted according to their wrench error prediction with $\texttt{Weight-Function}$ defined as $\vec{w}_t^i = exp(- \gamma ( \vec{\hat{f}}_{t}^i - \vec{f}_t) )$ and resampled with probability proportional to this weight vector.  Here, $\vec{\hat{f}_{t}^i}$ is the reaction wrench prediction of the $i^{th}$ particle and $\vec{f_{t}}$ is a ground truth measurement. Finally, the particles are propagated through the \textcolor{Aquamarine}{Force} and \textcolor{violet}{Action Modules}, a small amount of uncertainties is applied $\sim \mathcal{N}(0,0.01)$ to the particles and the algorithm is repeated. (see algorithm variations in \textbf{Appendix} Sec.~\ref{sec: Inference algorithm variations})



\begin{algorithm}
  \caption{Particle Filter based Inference Algorithm}\label{Alg: particle filter}
  \begin{algorithmic}[1]
    \Procedure{}{}
        \State $\mathbf { \hat{C}}_0 \gets \{ \vec{ \hat{c}}_0^0, \vec{ \hat{c}}_0^1, ..., \vec{ \hat{c}}_0^{n-1}\} \sim \mathcal{N}(0, 0.01)$
            \Comment{Init Particles}

        \For{each $scan$ in $Trajectory~Length$ }
        
            \State $\mathbf C_t \gets \texttt{Refine}(\mathbf{P}_t,\mathbf {\hat{C}}_t, \vec{f}_t, \vec{p}_t)$

                   
                \State $\vec{w}_t \gets \texttt{Weight-Function}(\vec{f}_t,\vec{\hat{f}}_{t}^0, \vec{\hat{f}}_{t}^1, ..., \vec{\hat{f}}_{t}^{n-1})$

                \State   $\mathbf{\bar{C}_{t}} \gets  \texttt{Re-sampling}(\vec{w}_t, \mathbf {C}_{t})$
            
            \State $\mathbf{ Z_t} \gets \texttt{Force-Module}(\mathbf {\bar{C}}_t$, $\mathbf f_t$, $\mathbf p_t, \mathbf{ \alpha})$    
            \State $ \{ (  \vec{\hat{f}}_{t+1}^0, \vec{\hat{c}}_{t+1}^0 ), ( \vec{\hat{f}}_{t+1}^1, \vec{ \hat{c}}_{t+1}^1 ), ..., ( \mathbf{\hat{f}}_{t+1}^{n-1}, \vec{\hat c}_{t+1}^{n-1} )\}  \gets$ \text{Action-Module} $( \mathbf{ Z_t}, \vec{ \alpha}, \vec a_t )$ 
            \State $t \gets t+1$
                
        \EndFor
        
    \EndProcedure 
  \end{algorithmic}
\end{algorithm}


\vspace{-5pt}
\section{Experiments \label{sec: Experiments}}
\vspace{-5pt}

For real robot experiments, we investigate two classes of objects: \textbf{spatulas} and \textbf{bike chains}. These objects exhibit different deformation behaviors: (i) spatulas are representative of bendable elastic objects that retain their shape when external forces are removed, whereas (ii) bike chains (with the 2 end points controlled by a bi-manual manipulator) can undergo significantly larger deformations (\eg from gravity) -- requiring algorithms that can reason over free-space while estimating in-contact shapes and forces. Spatula deformations are generally local and experience significant reaction wrenches, while chain deformations are much larger but with much smaller reaction wrenches.

\vspace{-5pt}
\subsection{Spatula Manipulation Dataset \label{sec: experiments - spatula scraping}}
\vspace{-5pt}

\begin{wrapfigure}{r}{0.2\textwidth}
  \vspace{-10pt}
 \centering
    \includegraphics[width=0.2\textwidth]{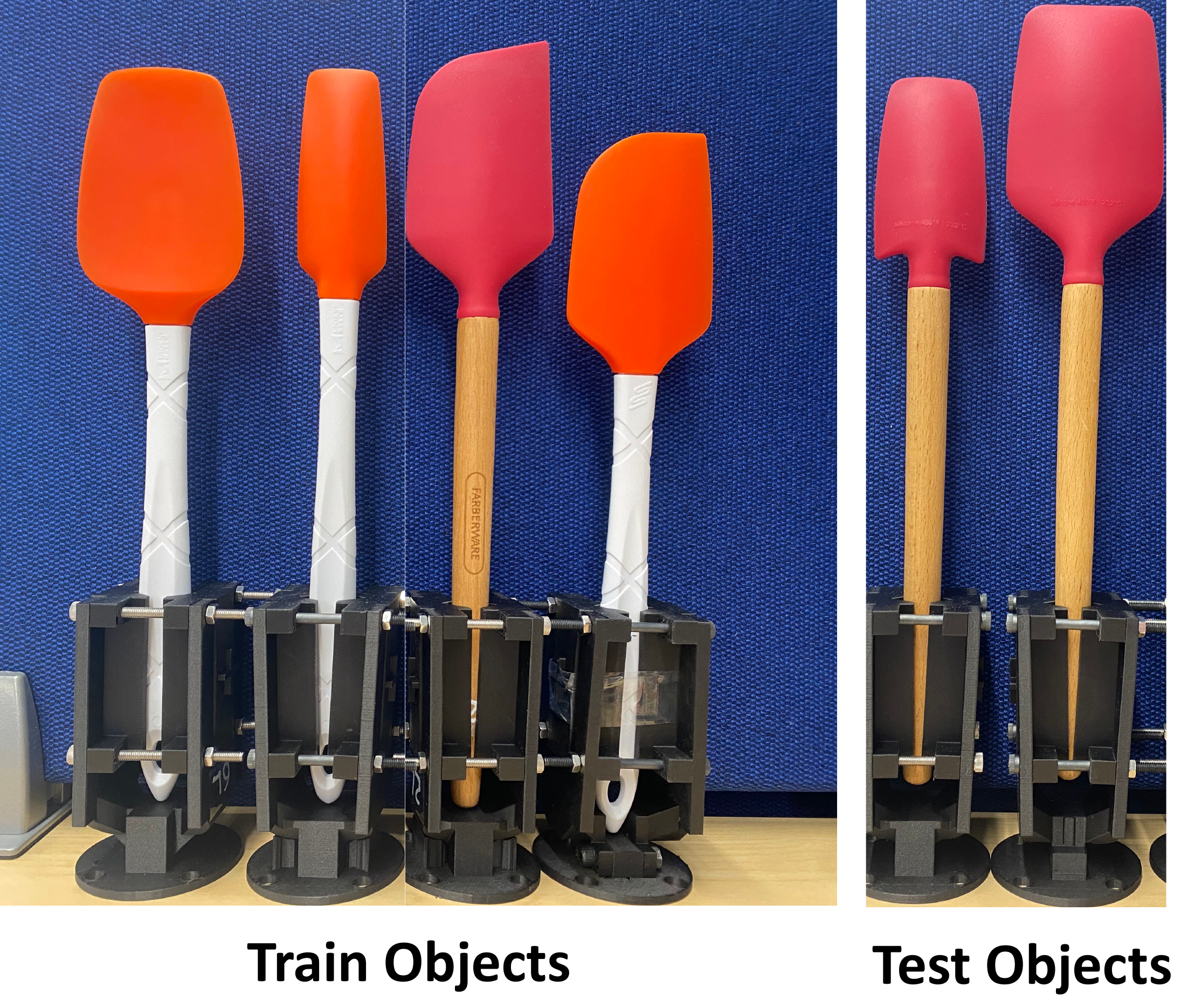}
      \caption{\small Train and test objects differ in terms of (i) geometry, and (ii) how they react under wrench.}
      \label{fig:spatula-armory}
  \vspace{-10pt}
\end{wrapfigure}

Train and test objects are shown in Fig.~\ref{fig:spatula-armory}. The training set consists of 4 spatulas, each with 672 samples ($N=28$, $k=24$), using 9:1 train/test split (25 train and 3 test trajectories) for representation learning. The test dataset consists of 2 held-out spatulas, each with 72 samples ($N=3$, $k=24$), and is used to evaluate zero-shot transfer and not used for representation learning. At the beginning of each trajectory ($k=0$), the end-effector moves to a random world Cartesian position in a bounding box of size $(0.1\times0.24\times0.06)$ meters and the tool is brought into contact with the table. Within a trajectory, the tool maintains contact while sampling a random action from $(x,y,z,r,p,y) = \pm(0.02, 0.02, 0.01, 0.05, 0.04)$ where translations are in meters and rotations in radians. Wrench measurements are averaged over 0.05 [s] and the point cloud is registered to the wrist frame. \rb{The method only requires a full point cloud from the nominal geometry of the object. This is collected by holding the object in front of the camera and rotating. 
Deformed object geometries are collected from a single camera viewing the front side of the spatulas without occlusions and are all partial. 
} $\vec P$ is then the result of segmenting the object pointcloud and normalizing the scale into a bounding box of size $2\times2\times2 [m]$ centered around $(0,0,0)$. The test dataset point clouds $\vec P$ contain occlusions at the bottom $0.15 \times \texttt{spatula-height}$ as shown in the top observation panel in Fig.~\ref{fig:teaser}. For hardware, we use Franka Emika's Panda, wrist mounted ATI Gamma F/T sensor, and Photoneo PhoXi 3D L. 

\vspace{-5pt}
\subsection{State Estimation and Predictions } \label{sec:State Estimation and Predictions }
\vspace{-5pt}

\begin{wraptable}[13]{r}{7cm}
\vspace{-10pt}
    \centering
        \begin{tabular}{lcc}\toprule
        CD \rb{$(\times 10^3)$} & Spatula w/ table &  Bike Chain  \\ \midrule
        Train Est.  & 0.777 (0.215)  & 0.708 (0.331)  \\
        Train Pred.  &  0.830 (0.330) & 0.762 (0.224)  \\ \midrule
        Test Est.  & 0.934 (0.210)  & 0.760 (0.234) \\
        Test Pred.  & 1.041 (0.348)  & 1.247 (0.391) \\
        \bottomrule
        \end{tabular}
    \caption{\small \accro can accurately estimate object geometry deformations under significant occlusions, measured in terms of mean Chamfer Distances (CD) (std.) $[m^2]$ scaled as $(\times 10^3)$ across 5600 on-surface points between predicted and ground truth reconstructions.} 
    \label{tab: chamfer distance of both demo} 
\setlength{\textfloatsep}{2.5\baselineskip}
\end{wraptable}

\begin{figure}[thpb]
  \centering
     \vspace{-10pt}
    \includegraphics[width=1\textwidth]{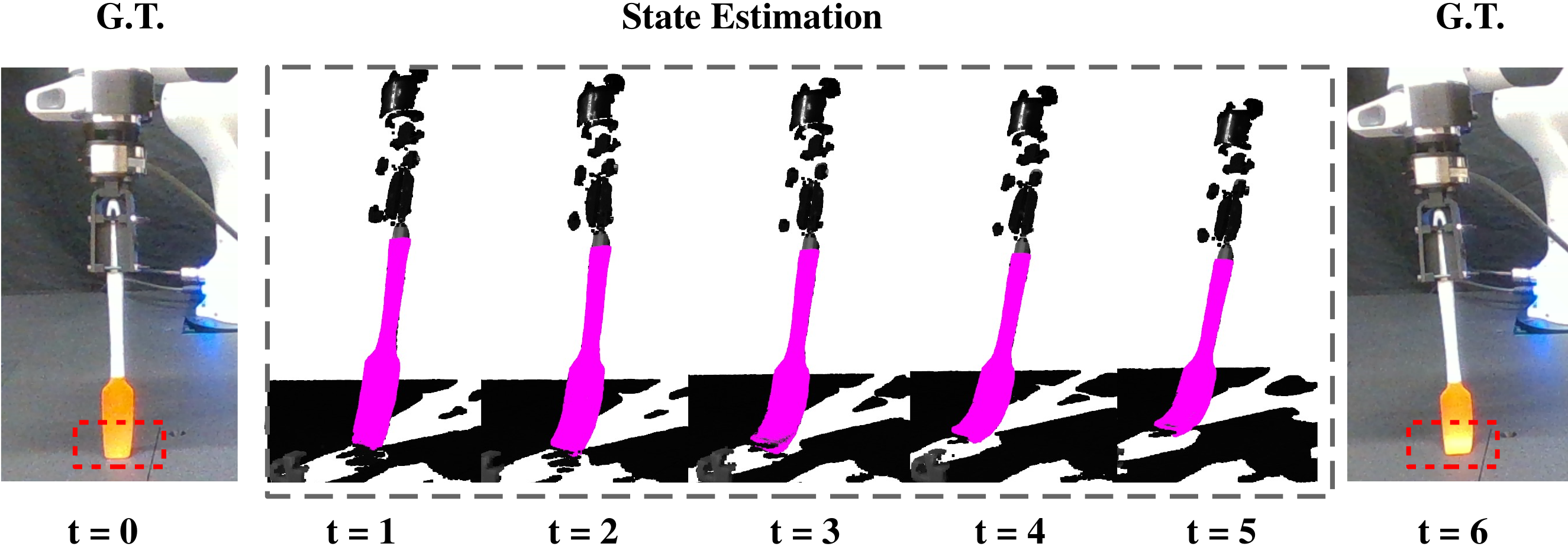}
    \caption{\small
    Even under occlusions (generated by cropping out data from the within the red dotted box), our approach can accurately infer the how the geometry of a test spatula deforms over time (t = 0 to 6), indicated by high overlap agreement between the prediction (magenta) overlaid on the ground truth (black).
    }
    \vspace{-10pt}
  \label{fig:spatula state estimation}
\end{figure}

Here, we evaluate the deformation prediction (Sec.~\ref{sec: def-dynamics}) and state-estimation (Alg.~\ref{Alg: particle filter}) accuracy of \accro on training and test trajectories. Fig.~\ref{fig:spatula state estimation} shows an example of state-estimation for a test trajectory where the ground truth (black) and reconstruction (magenta) are overlaid. We note the high state-estimation accuracy indicated by the overlay agreement, where the average distance between the measured points and reconstructions is within 0.03$[m]$ of normalized scale, just 1\% of the object length. The quantitative results, Tab.~\ref{tab: chamfer distance of both demo}, show \accro is able to faithfully predict and estimate object deformations and reaction forces despite significant occlusions. Force and torque norm errors are less than 0.567 [N], 0.106 [Nm] (train) and  1.130 [N], 0.135 [Nm] (test) respectively (see \textbf{Appendix} Sec.~\ref{sec: Distribution of geometry and wrench state estimation and predictions} and Sec.~\ref{appendix - reaction wrench prediction errors} for details).


Our ablation study, Tab.~\ref{tab: ablation study}, helps further anchor the scale of these state-estimation and dynamics prediction errors. On state estimation, for example, our full VIRDO++ method achieves a Chamfer distance ($\times 10^3$) of normalized geometry is 0.93 $[m^2]$. \rb{The baseline model without a novel \cntemb~ input has 27\% increased chamfer distance error as 1.18 $[m^2]$ for geometry estimation. Without $c_t$, the baseline does not reason about contact during inference, resulting in larger errors during state-estimation} Further if we assume the object is rigid and reduce our state estimation to a ``rigid pose estimation'' pipeline, the Chamfer distance ($\times 10^3$) increases considerably to 16.19 $[m^2]$. 


\rb{The geometry predictions are slightly less accurate than state-estimation because the geometry prediction are computed by the learned dynamics using previous observation and action, while state-estimation uses the current measurement to update the prediction.} Despite this, the prediction Chamfer distances are within 14\% of the state estimation results. Moreover, the train and test trajectories results show 25\% difference in Chamfer distance. This is interesting because the test trajectories are not only from unseen tool configurations but also with occlusions (see \textbf{Dataset}). This suggests that learned dynamics of \accro generalize effectively to unseen contact formations with occlusions. 

\vspace{-5pt}
\subsection{Evaluation on a Downstream Manipulation Task - Extrinsic Contact Detection \label{sec: contact location detection}} 
\vspace{-5pt}
Central to compliant tool manipulation is estimating and controlling the contact between the tool and environment. For example, when scraping a wok with a spatula, the robot must reason over the contact formation (e.g., point, line, or patch) to ensure the spatula is scraping properly at the desired region on the wok. In this section, we evaluate \accro ability to estimate extrinsic contact features; specifically, the contact line created as the spatula is brought into contact with the table. To this end, we compute the signed distance values of the tables surface $(z=0)$ to compute a contact cluster. 
This requires only a single feed-forward path where the input is an $N\times3$ matrix, where N is the number of query points. From the contact cluster, we extract a contact line by selecting the two points furthest apart similar to \cite{ma2021extrinsic, qiang2013control}. The evaluation metrics are the average Euclidean distance 
$\frac{1}{2}\Sigma_{i=1}^{2}||l_i - l_i^* ||$, where $l_i$ is the detected contact lines, $l_i^*$ is the ground truth contact lines, and $i = 0, 1$ refers to right and left ends of each contact line respectively. Ground truth contact lines were obtained from ground truth pointcloud without occlusions by applying a contact mask $ z \leq \epsilon$. This contact feature estimation task is particularly interesting as it is 
the inverse problem of \cite{young2022virdo}; estimating \cite{young2022virdo}'s privileged input (=contact location) by inferring $c_t$. 

\begin{figure}
  \centering
    \includegraphics[width=1\textwidth]{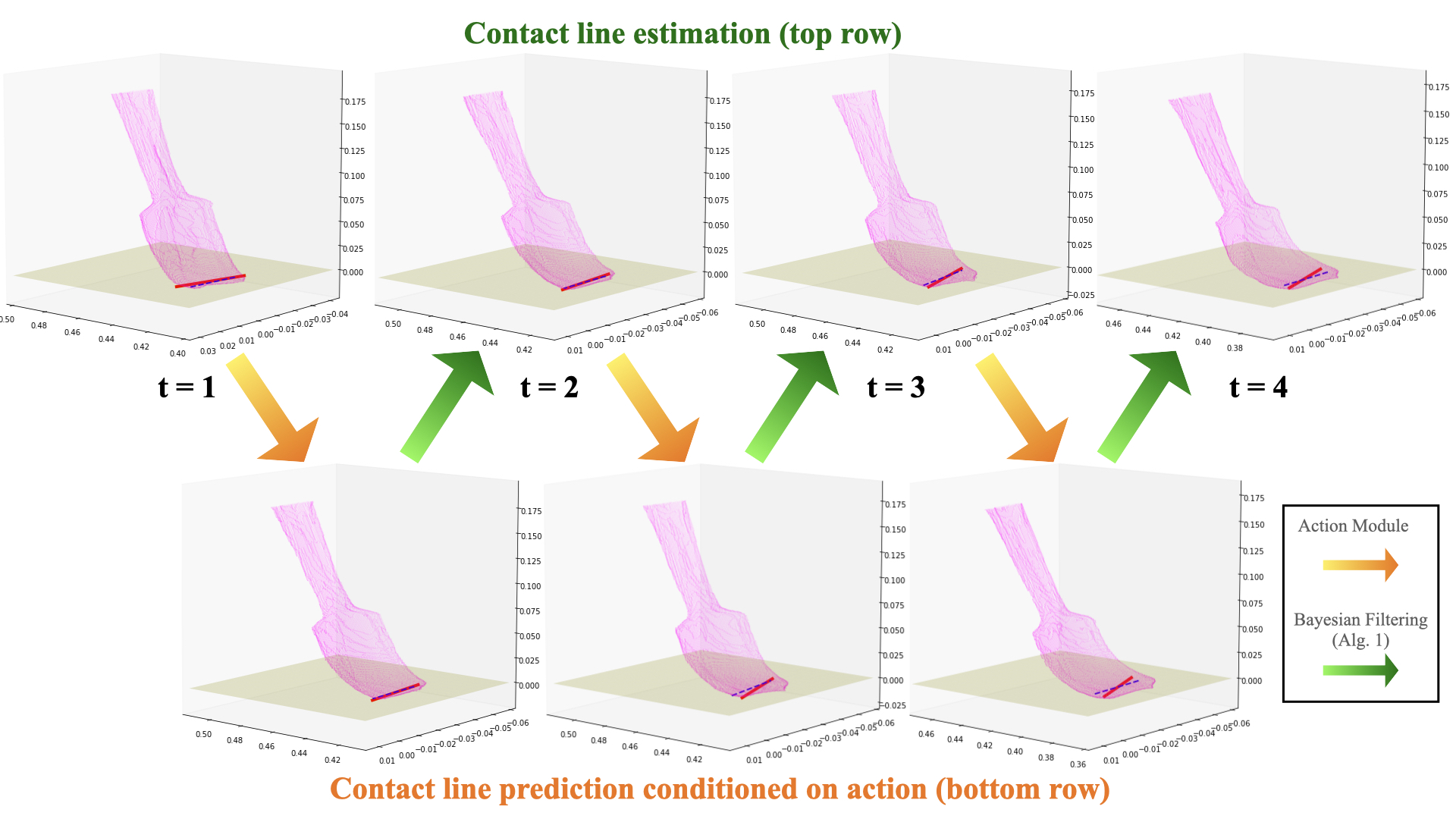}
    
    \caption{\small{\accro detects the line of contact throughout an alternating sequence of ``estimation $\xrightarrow{}$ prediction $\xrightarrow{}$ estimation $\xrightarrow{}$  ...''. Bayesian filtering increase the accuracy of contact location detection as the red lines (detected contact line) gets closer to the dotted blue lines (ground truth) following the green arrow. In each figure, yellow plane: table location, magenta spatula:  reconstructed geometry from \accro, red line: detected contact line prediction, blue: ground truth contact line. All axis are in [m] scale in world coordinate.}}
  \label{fig: train and test spatulas bend differently the same pressing force}
\end{figure}

\rb{As shown in Tab.~\ref{tab: ablation study on contact location estimation}, \accro can accurately detect contact locations with less than 8 mm error on a flat table for the current and the following time step regardless of occlusions at the tip. As in  Sec.\ref{sec:State Estimation and Predictions }, test trajectories from 4 spatulas with 15\% occlusion were used for the evaluation. Tab.\ref{tab: ablation study on contact location estimation} shows that the average contact line detection errors from all 4 spatulas are less than 7 [mm] during estimation and are less than 8 [mm] for predictions in Euclidean error. Similar to the result in Sec.~\ref{sec:State Estimation and Predictions }, estimations are 1 [mm] more precise than predictions because they are optimized using the sensor measurements. 
Considering that the total length of the spatulas are about 200 [mm], this is less than 4\% of error in total length. Estimating extrinsic contacts is a challenging task for rigid object \cite{sipos2022simultaneous} and is further complicated by object compliance; however, our approach proves effective due to its high-quality 3D reconstructions (details of error distribution in \textbf{Appendix} Fig.~\ref{fig: contact location detection }).} 
Tab.~\ref{tab: ablation study on contact location estimation} also quantifies performance against ablations (without $c_t$ and Rigid-body assumption) illustrating the importance of the contact embedding and reasoning over compliance.


\begin{table}
    \centering
        \begin{tabular}{lcccc}\toprule
        L2 Error [mm] & \accro & w/o $c_t$ & Rigid body asmp. \\ \midrule
        Contact Est.  & 6.647 (4.806) &  9.872 (5.725) & 21.203 (7.031) \\
        Contact Pred.  & 7.198 (4.439) & 9.356 (5.978)  &  22.572 (6.990)\\
        \bottomrule
        \end{tabular}
    \caption{\small Having latent embedding $c_t$ as an input, \accro can decrease the about 33 \% of error for estimations and 23 \% for predictions. Rigid body assumption is a worst case scenario producing $\times 3$ error. 
    } 
    \label{tab: ablation study on contact location estimation} 
     \vspace{-20pt}
\end{table}

\vspace{-5pt}
\subsection{Generalizations Tasks}
\vspace{-5pt}

This section demonstrates \accro generalization ability to unseen objects and their dynamics. For each unseen object, the generalization dataset contains 1 nominal point cloud and 3 trajectories, each with 24 transitions with occlusion. Here, we directly apply the same pre-trained model from Sec. \ref{sec: experiments - spatula scraping} to the generalization test dataset.

\vspace{-5pt}
\subsubsection{Object code inference for Unseen Objects\label{sec: Object code inference}} 
\vspace{-5pt}

\begin{wrapfigure}[13]{r}{7cm}
    \vspace{-15pt}
        \includegraphics[width=7cm]{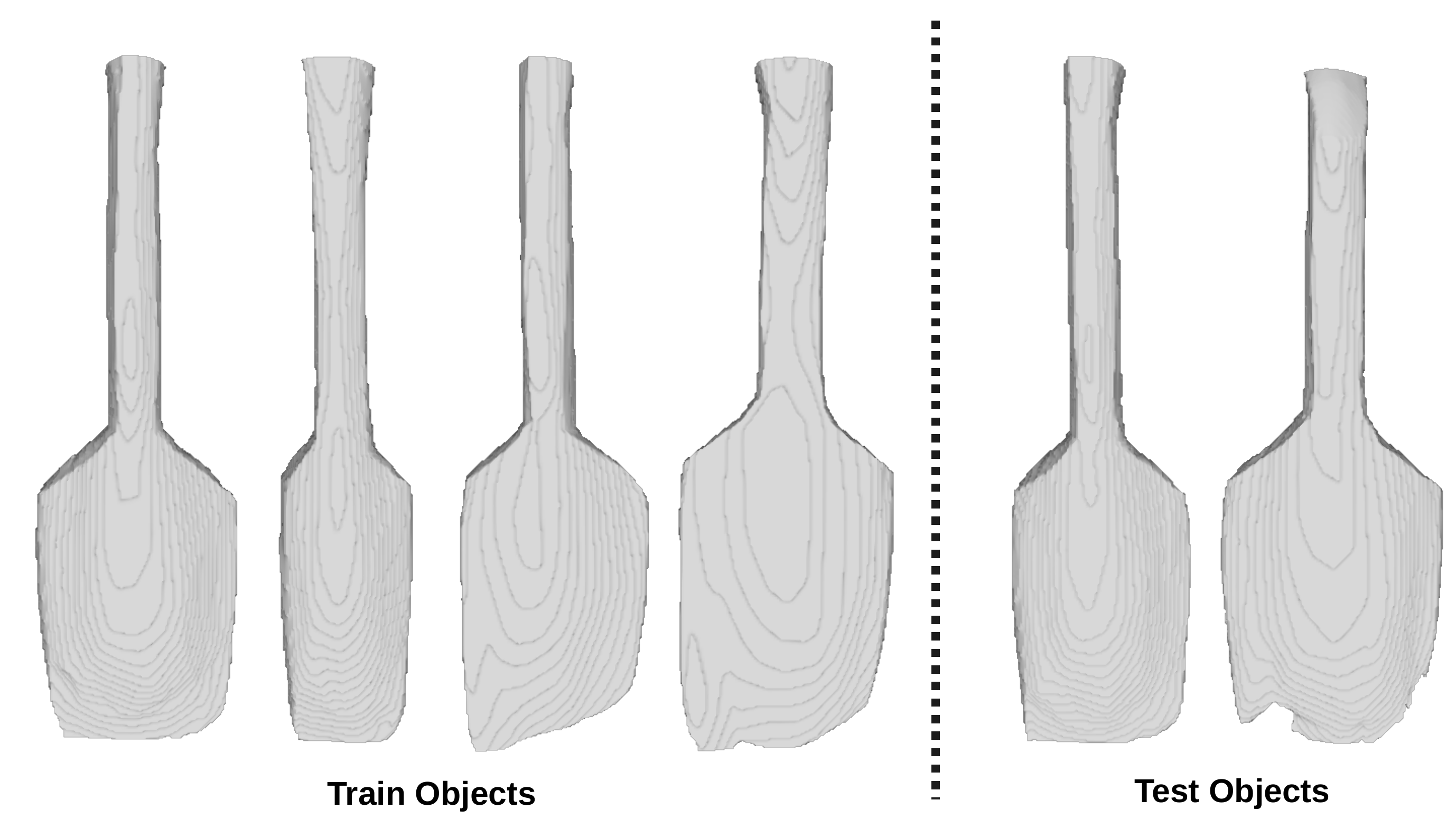}
    \caption{\small\accro can reconstruct train and test objects with high fidelity (same order as Fig.~\ref{fig:spatula-armory}).}
    \label{fig: object module generalization}
\end{wrapfigure}

\accro is able to represent nominal geometries unseen during training within distribution. Fig.~\ref{fig: object module generalization} shows reconstructions of both seen and unseen object in Fig.~\ref{fig:spatula-armory} using a single pretrained Object Module. For the unseen objects, we start from a randomly initialized object code $\alpha \sim \mathcal{N}(0, 0.01)$ and update the code by gradient descent with Adam. We use the same loss function from Eq.~\ref{eq: pretrain loss} with pre-trained \accro while freezing the weights of the network. The first row of Tab.~\ref{tab: generalization} shows the Chamfer distance error of the unseen objects alongside training objects, indicating the relatively small drop in performance. When comparing Fig.~\ref{fig: object module generalization} and the ground truth in Fig.~\ref{fig:spatula-armory}, we note that the test reconstructions are close to the ground truth aside from some fine edge details.

\vspace{-5pt}
\subsubsection{Zero Shot Dynamics Prediction and State-estimation Generalization} 
\vspace{-5pt}

With the object codes obtained in Sec.~\ref{sec: Object code inference}, \accro is able to predict the unseen object's dynamics based on their nominal shape. Here, we evaluate both prediction and state-estimation performance as in Sec.~\ref{sec:State Estimation and Predictions }. We note that the observations of the unseen objects include occlusions as discussed in \textbf{Dataset}. Tab.~\ref{tab: generalization} shows comparable performance between training and test objects with a small drop in accuracy, highlighting our model's generalization capability. Sec.~\ref{sec: details on generalization to novel objects dynamics} in \textbf{Appendix} provides intuition for the Chamfer distance values. Further, we evaluate \accro's ability to perform high-fidelity geometry prediction and estimation even in novel environment (Wok scraping) without 3D model. Details can be found in \textbf{Appendix} Sec.~\ref{appendix - generalization to novel environments}.

\begin{table}[!htb]
    \centering
    \small
        \begin{tabular}{l|ccc|cc }\toprule
             CD ($\times 10^3$) & \accro  &  w/o $\vec c_t$ & Rigid Asm.  & Test Obj. 1 & Test Obj. 2 \\ \midrule
            Geo. -- Nom. & 0.65 (0.08) &  - & -   & 1.08  & 1.29  \\
            Geo. -- S.E.   &  0.93 (0.21) & 1.18 (0.35) & 16.66 (11.51) & 1.39 (0.20)  & 1.56 (0.29)\\
            Geo. -- Pred. &  1.04 (0.35) & 1.09 (0.25) &  15.16 (10.57)  & 1.54 (0.45)  & 2.40 (1.08) \\
            Wr. -- Pred. & 1.14 (1.05) & 1.04 (1.01) & 1.63 (0.89)  & 1.50 (1.00) & 2.14 (1.75)  \\ \bottomrule
        \end{tabular}
    \caption{\small \rb{(column 1-3) Ablation study using the train object's unseen trajectories. Numbers indicate mean (std.) of each error distribution. Chamfer distance used 5600 points and is multiplied by $10^3$. (column 4-5) Generalization results to unseen objects using the same pretrained \accro. Geo. = Geometry, Nom. = Nominal, S.E. = State Estimation, Pred. = Prediction, Wr. = Wrench calculated as  $ \| \vec{ \hat {f}}_t -\vec{f}_t \|$.} 
    \label{tab: ablation study} \label{tab: generalization}}
    \vspace{-10pt}
\end{table}



\vspace{-5pt}
\subsection{Bike Chain Application \label{sec: bike chain demo}}
\vspace{-5pt}

\begin{figure}
  \centering
    \includegraphics[width=1\textwidth]{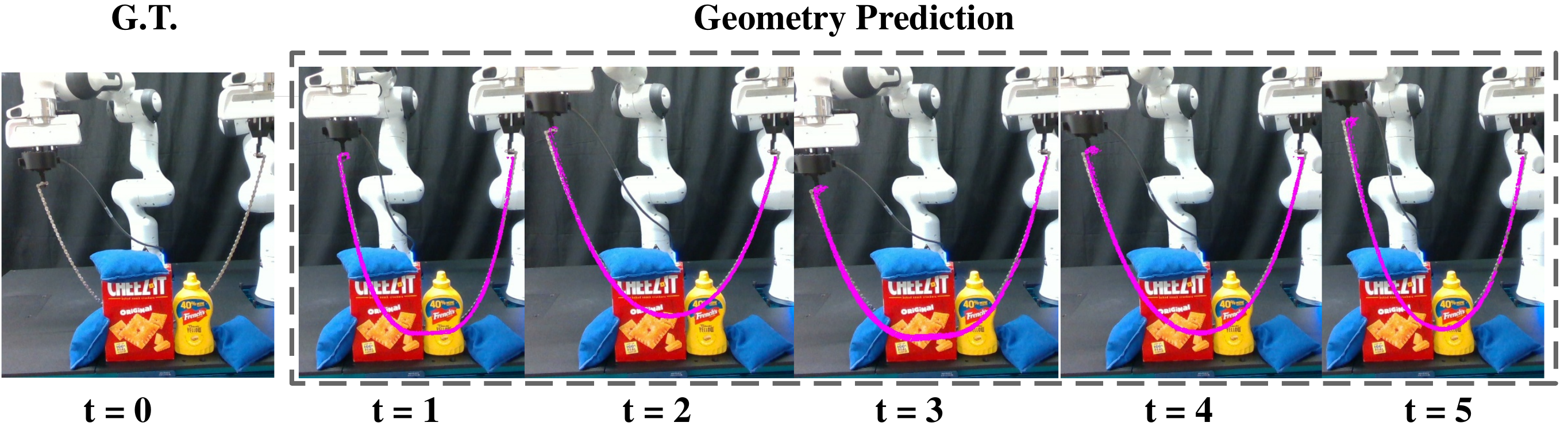}
    \vspace{-15pt}
    \caption{\small
    \accro can accurately estimate the bike chain geometry (predictions overlaid in magenta) under occlusions due to objects at the front. Note that the chain intermittently makes contact and rests on the table, which is correctly predicted by \accro despite significant occlusion of where the chain makes contact.
    }
    \vspace{-10pt}
  \label{fig:bikechain}
\end{figure}

In this experiment, the chain ends are grasped by two arms. One end of the chain is moved randomly through the vertices of an equally-spaced $6\times 6\times 6$ grid fitting inside of a $0.1 \times 0.18 \times 0.2$ meter bounding box. The visited vertices of this grid make up the training dataset. The test dataset is generated by sampling 20 random positions uniformly from the bounding box. 
The end-effector orientations were fixed; 
therefore, $\vec{p}_t, \vec{a}_t \in \mathbb{R}^{3}$. As shown in Fig.~\ref{fig:bikechain}, \accro is able to predict/estimate both free-space and in-contact chain geometry. The in-contact phase occurs when the arm moves sufficiently far down such that the chain lays on the table. We report the quantitative results of geometry estimation and dynamics prediction in Tab.~\ref{tab: chamfer distance of both demo} and \textbf{Appendix}  Sec.~\ref{sec: Distribution of geometry and wrench state estimation and predictions}. The empirical results are comparable to spatula experiments despite significantly larger shape variations. Details on wrench prediction errors in \textbf{Appendix} Sec.~\ref{appendix - reaction wrench prediction errors}.

\vspace{-5pt}
\section{Limitations} 
\vspace{-5pt}

Our proposed approach has a number of limitations:
\textbf{1) Non-rigid grasps:} \accro assumes that the gripper is rigidly holding the object. When it comes to non-rigid grasping, we need to estimate relative motion of the objects w.r.t the gripper and the force transmission is now a function of this relative motion. One potential solution is to use collocated tactile sensor feedback \cite{taylor2021gelslim3, yin2022multimodal, kuppuswamy2020soft} with F/T sensing to reason over reaction forces conditioned on the grasp. \textbf{2) Inertial effects:} our approach assumes quasi-static manipulation (negligible accelerations). During fast motions, the tools' inertial effects may also impact deformations and wrench measurements. When applicable, we suggest supplementing \accro's input with end-effector accelerations. \textbf{3) Generalization to novel environments:} While we provide an example of environment generalization (wok experiments Fig.~\ref{fig:wok-scraping}), the performance is lower than our other demonstrated applications. \accro may benefit from a 
3D model of the environment to integrate inter-penetration losses \cite{petit2018capturing} or collision checkers \cite{mitrano2021learning}.





\clearpage
\acknowledgments{This work was in part supported by Robotics at Google under the Google Faculty Research Award 2021 and the National Science Foundation (NSF) grant NRI-2220876. Any opinions, findings, and conclusions or recommendations expressed in this material are those of the authors and do not necessarily reflect the views of the National Science Foundation.}


\footnotesize
\bibliography{reference}  

\newpage

\appendix
\section{Appendix}

\localtableofcontents

\subsection{Training Details}
Tab.~\ref{tab: training details} contains training parameters. In addition, we used 2,000 on-surface points and 20,000 off-surfaces for training \accro. We used the Adam optimizer with learning rate = $10^{-4}$ and epoch = 200.

\begin{table}[h]
    \centering
        \begin{tabular}{lccccccccccccc }\toprule
         $l_c$ & $\lambda$ & $\lambda_1$ & $\lambda_2$  & $\lambda_3$ &  $\lambda_4$   & $\lambda_5$ &  $\lambda_6$&  $\lambda_7$ &  $\lambda_8$ & $\lambda_9$ & $\lambda_{10}$ &  $\lambda_{11}$ \\ \midrule
         6 & 5e1 & 1e1 & 1e1 & 1e4 & 1 & 1e2 &  1e5 & 5e1 & 3e6 & 1e1 & 1e1 & 1e1 \\ \bottomrule

        \end{tabular}
    \caption{ \label{tab: training details} Parameters used for training \accro}
\end{table}

\subsection{Inference Details}
\subsubsection{Parameters \& Hardware}
For the inference, we use number of particles $n = 40$, resampling parameter $\gamma = 0.7$, and epoch $e = 10$. This results in inference time = 0.7 [s] with GPU = NVIDIA A6000 and CPU= 32-Core 3.50 GHz AMD. We note that one can balance the state-estimation accuracy between the inference time by changing the number of particles and number of epochs. Here, we note that the epochs need not to be large due to Alg.~\ref{Alg: particle filter} which provides a close-to-ground-truth initial guess on the object's state.

\subsubsection{Inference Algorithm Variations \label{sec: Inference algorithm variations}}

\begin{algorithm}
  \caption{Resampling Algorithm}\label{Alg: particle filter - resampling}
  \begin{algorithmic}[1]
    \Function{Re-sampling}{$\vec{w}_t, \mathbf {C}_{t}$}
        \State Draw  $\{ \vec{\bar{C}}_{t}^{0}, \vec{\bar{C}}_{t}^{1}, ..., \vec{\bar{C}}_{t}^{k-1} \} \sim \mathcal{N}(0,0.01)$ 
        \Comment{Exploration}
        \State Draw  $\{ \vec{\bar{C}}_t^{k}, \vec{\bar{C}}_t^{k+1}, ..., \vec{\bar{C}}_{t}^{N-1} \} \sim \{ (\vec{C}_{t}^{0}, w^0_t),(\vec{C}_{t}^{1}, w^1_t) ..., (\vec{C}_{t}^{N-1}, w^{N-1}_t) \}$ 
        \Comment{Exploitation}
        
    \State \Return{$\mathbf{\mathbf{C}}_{t}$}
    \EndFunction 
  \end{algorithmic}
\end{algorithm}

The input of the resampling function in Alg.~\ref{Alg: particle filter} is the beliefs over the current states ($\mathbf{C}_t$) and the corresponding reliability, a.k.a the sampling weight ($\vec{w}_t$). The resampling function refines $\mathbf{C}_t$ to $\mathbf{\bar{C}}_t$ through Alg.~\ref{Alg: particle filter - resampling}. In the exploitation step, we compose $k$ particles in $\mathbf{\bar{C}}_t$ by sampling from $\sim \mathcal{N}(0,0.01)$. In the exploration step, we sample the rest of $\mathbf{\bar{C}}_t$ by sampling from $\vec w_t$. 

Our ablation study (presented in Tab.~\ref{Tab: particle filter - resampling}) aims to find an optimal balance between exploration and exploitation. Here, the ``Force-Prediction'' is calculated as $\|\hat{\vec f}_t[:3] - \vec f_t[:3] \|$ and ``Torque-Prediction'' is calculated as $\|\hat{\vec f}_t[3:] - \vec f_t[3:] \|$. 
We found that $\beta = 0$ and $25\%$ has comparable state estimation results. Geometry prediction is the most accurate when $\beta=0\%$; however, wrench predictions are most accurate when $\beta=25\%$. When compared to $\beta=0\%$, $\beta=25\%$ has a $10 \%$ increased geometry prediction error and $9-10\%$ reduced wrench prediction error.

\begin{table}
    \centering
        \begin{tabular}{lccccc }\toprule
         CD ($x10^3$) & $\beta = 0$\% & $\beta = 25$\%  & $\beta =50$\% &  $\beta = 75$\%  \\ \midrule
        Geom. -- S.E. &  0.936 (0.206)& 0.937 (0.209) & 0.941 (0.236) & 0.943 (0.202)  \\
        Geom. -- Pred.  &  1.035 (0.335)& 1.046 (0.358) & 1.049 (0.370) & 1.117 (0.497) \\ 
        Force -- Pred. &  1.241 (1.282))& 1.125 (1.068)  &  1.149 (1.069) & 1.326 (1.282)\\ 
        Torque -- Pred. &   0.140 (0.109)&0.134 (0.099) & 0.134 (0.095) & 0.158 (0.111) \\ \bottomrule
        \end{tabular}
    \caption{ \small $\beta=0 \sim 50 \%$ does not significantly increase nor decrease the performance of our Inference algorithm. However, when exploration becomes dominant over the exploitation ($\beta = 75\%$), a distinguishable drop in performance is observed. \label{Tab: particle filter - resampling}}
\end{table}

\subsection{Experiment Details}

\subsubsection{Differences in Physical Properties of Spatulas}
\begin{figure}
  \centering
    \includegraphics[width=0.6\textwidth]{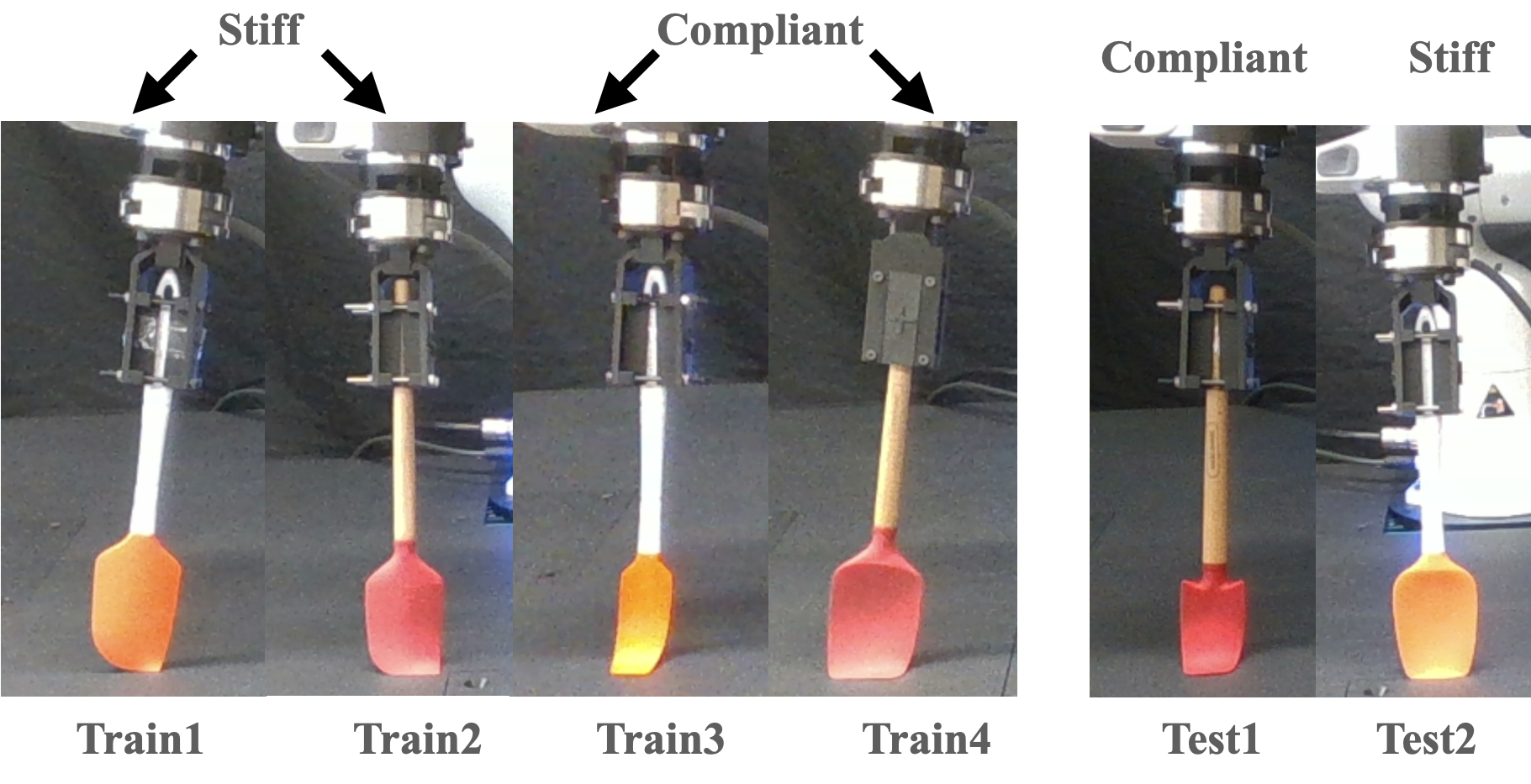}
    \caption{\small We maintained the pressing force of 6 [N] for all six spatulas and measured the difference in deformations. The combination of material property and geometry determines stiffness/compliance. The spatulas with the same tip color consist of the same material; however, they deform very differently depending on their structures. \label{fig: Differences in Physical Properties of Spatulas }} 
\end{figure}

In this section, we illustrate how the train/test spatulas deform relative to each other when brought into contact with the table in a similar configurations (affected by the arm impedance controller) and same normal force (here, 6 [N] along the $z$ axis). Fig.~\ref{fig: Differences in Physical Properties of Spatulas } shows a wide distribution of deformations achieved within the same object category. With the same force, train object 1 barely deforms and is close to it's nominal shape, whereas train object 3 bends significantly. This difference in physical properties makes generalization very difficult, even within the same object class. For example, test object 2 visually resembles train object 4; however, it has stiffness properties more similar to train object 2. This has lead to \accro producing higher errors in the generalization task for test 2 objects.

\subsubsection{Coordinate System and Action Distribution}

\begin{figure}
  \centering
    \includegraphics[width=0.3\textwidth]{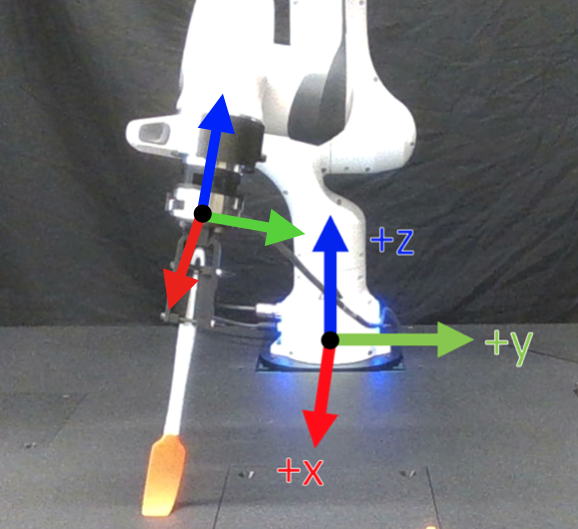}
    \caption{Action frame is defined at the robot's base frame while pointclouds are registered with respect to the end-effector frame.}
  \label{fig: action frame}
\end{figure}

\begin{figure}
  \centering
    \includegraphics[width=1\textwidth]{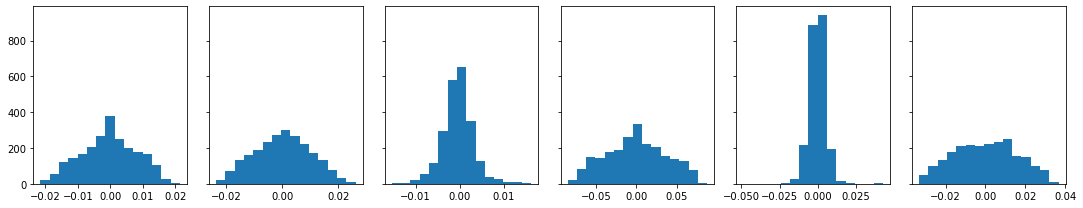}
    \includegraphics[width=1\textwidth]{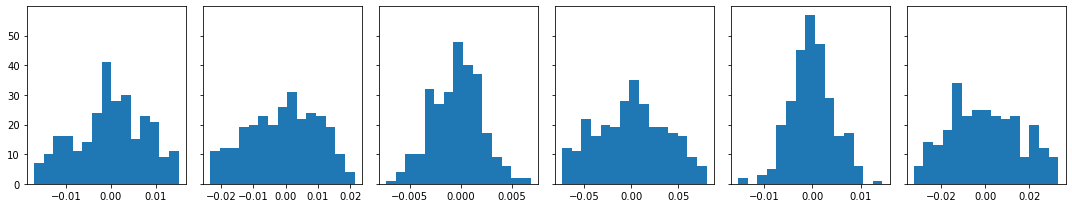}
    \caption{\small Distributions of training and testing action trajectories. These distributions are intentionally centered around zero. Top row = Train, Bottom row = Test. Fx, Fy, Fz, Tx, Ty, Tz order where x-axis is either force[N] or Torque[Nm]. y-axis quantifies the number of samples.}
  \label{fig: action distribution }
\end{figure}

Pointcloud are collected with respect to a reference frame defined at the end of the F/T sensor mounted at the robot's wrist. Actions are the Cartesian motion of the end-effector of the robot with respect to the base frame, Fig.~\ref{fig: action frame}. Fig.~\ref{fig: action distribution } visualizes action distributions of train and test trajectories of training objects. Both are normally distributed and share mean and standard deviation because they were generated from the same  action policy.

\subsubsection{Distribution of Geometry and Wrench State Estimation and Predictions} \label{sec: Distribution of geometry and wrench state estimation and predictions}

\begin{figure}[thpb]
  \centering
    \vspace{-5pt}
      \includegraphics[width=1\textwidth]{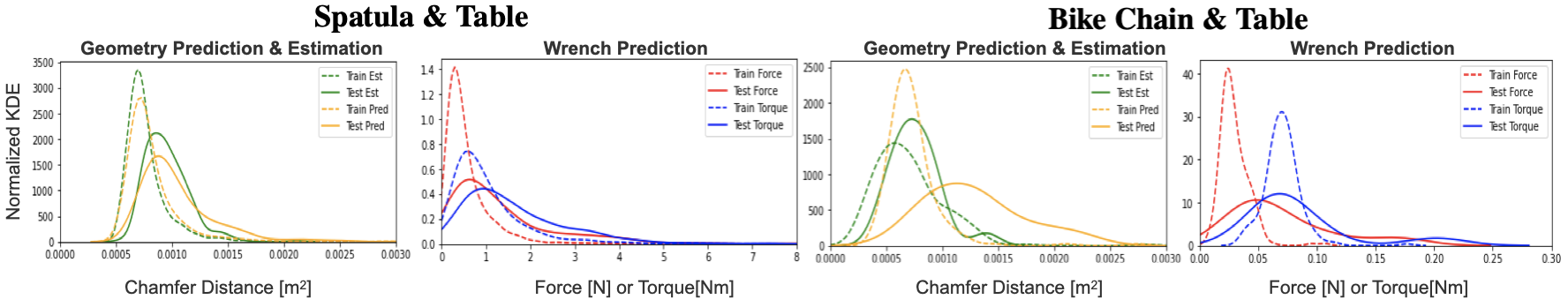}
    \caption{Kernel Density Estimate (KDE) plots for spatula and bike chain applications show that accro's geometry and wrench predictions and state-estimation errors are within small margins. Force prediction errors  = $\|\hat{\vec{f}}_t[:3]- \vec{f}_t[:3]\|$ [N]. Torque prediction error as $10*\| \hat{\vec{f}}_t[3:]- \vec{f}_t[3:]\|$ [Nm].}
  \label{fig: KDE Train and Test result of both demos}    
  \vspace{-10pt}
\end{figure}

Fig.~\ref{fig: KDE Train and Test result of both demos} shows detailed distribution of errors reported in Tab.~\ref{tab: chamfer distance of both demo}. The Kernel Density Estimate (KDE) plot shows that the errors are normally distributed. The training error mean is smaller than the test error for all graphs in both demos. In addition, geometry estimation is more accurate than geometry predictions as it uses sensor measurements. For example, in the spatula demo, the Chamfer distance error at the peak geometry estimation from the training set (green dashed line) is smaller than the test set (solid dashed line) and is also smaller than the prediction estimates (yellow dashed line).

\subsubsection{Reaction Wrench Prediction Errors \label{appendix - reaction wrench prediction errors}}
Tab.~\ref{tab: reaction force predictions on both demo} shows the wrench predictions results of trained objects (Tab.~\ref{tab: generalization}'s first column). Because we use L2 norm for reaction wrench prediction during training, error distributions are centered around 0, and the degree of errors appears as standard deviation. 

The standard deviation in error for the chain task is approximately an order of magnitude smaller than the spatula. We hypothesize that this because i) the ground truth wrench is an order of magnitude smaller than the spatula and ii) the action space dimension is half the spatula resulting in a simpler model to learn.

\begin{table}[ht]
    \centering
        \begin{tabular}{lcccccc}\toprule
         $l_1$ Error & $f_x$ [N]& $f_y$ [N]& $f_z$ [N]& $\tau_x$ [Nm] & $\tau_y$ [Nm] &  $\tau_z$ [Nm] \\ \midrule
         Sp. Train   &   0.01 (0.31) & -0.00 (0.36) &  0.03 (0.65) & -0.00 (0.11) & 0.00 (0.09) &  0.00 (0.02)\\
        Sp. Test  & -0.03 (0.35) &  0.08 (0.54) & -0.28 (1.60) & -0.02 (0.16) & -0.01 (0.11) & -0.00 (0.02)  \\
        Ch. Train & 0.02 (0.01) & 0.01 (0.01) &  0.02 (0.01) & -0.01 (0.00) & 0.00 (0.01) & 0.01 (0.00) \\
        Ch. Test & 0.02 (0.06) &  0.01 (0.03) & 0.03 (0.01) & -0.00 (0.00) & 0.00 (0.01) &  0.00 (0.00) \\ \bottomrule
        \end{tabular}
    \caption{\small\textbf{Mean and Std Wrench Prediction Errors:} $ \vec{ \hat {f}}_t -\vec{f}_t$ 
    for the spatula and chain experiments.
    \label{tab: reaction force predictions on both demo}
}
\vspace{-10pt}
\end{table}
 
\subsubsection{Details o Contact Detection Errors}

\begin{figure}
  \centering
    \includegraphics[width=0.6\textwidth]{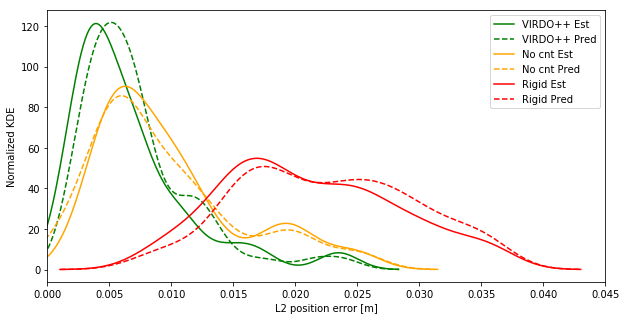}
    \caption{\accro's contact detection outperforms the baselines without $c_t$ and rigid body assumption where the detailed quantitative analysis is in Tab.~\ref{tab: ablation study on contact location estimation}  \label{fig: contact location detection }}
   \vspace{-10pt}
\end{figure}

Fig.~\ref{fig: contact location detection } contains detailed error distributions of Tab.~\ref{tab: ablation study on contact location estimation} from the main text. \accro (green)'s Chamfer distance has a significantly smaller mean and standard deviation compared to the baseline models without contact embedding and compliance assumption. One interesting observation is that the geometry prediction (yellow dashed line) is located left to the geometry estimation (solid yellow line). This is because the model without $c_t$ produces deterministic reconstructions as a function of $p_t$ and $f_t$, leaving no room for improving state estimation with the Bayesian filtering algorithm.

\subsubsection{Details on Generalization to Novel Objects Dynamics \label{sec: details on generalization to novel objects dynamics}} 

\begin{figure}[thpb]
  \centering
    \vspace{-8pt}
    \includegraphics[width=0.85\textwidth]{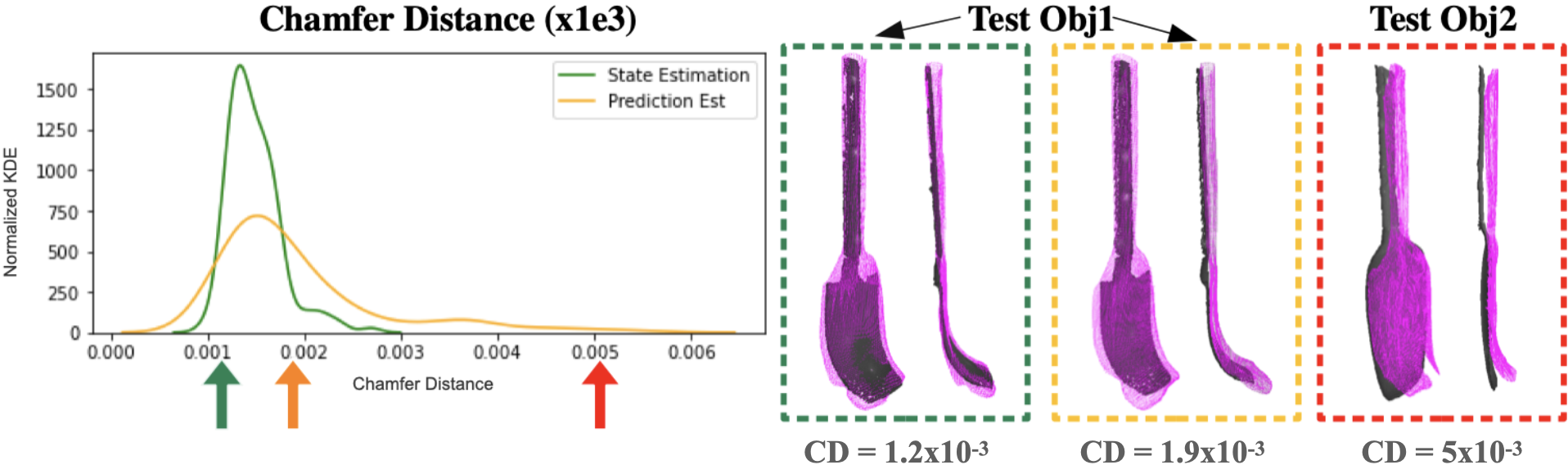}
    \vspace{-5pt}
    \caption{\small\textbf{Generalizing to unseen object dynamics}. Left panel: KDE plot of Chamfer distances (x1e3) [$m^2$] using normalized geometries. We show both test objects state estimation using partial point clouds and geometry predictions. Similar to training objects, state estimation has smaller mean and standard deviation of the error. Right panel: We visualized three representative generalization results and corresponding Chamfer distances. \label{fig: dynamics generalization to novel objects}}
    \vspace{-10pt} 
\end{figure}

Fig.~\ref{fig: dynamics generalization to novel objects} is a detailed Chamfer distance distribution of Tab.~\ref{tab: generalization} from the main text and the visualization of each Chamfer distance loss. The purpose of reconstructions visualized in Fig.~\ref{fig: dynamics generalization to novel objects} (right) is to further provide an intuition of each error and how to interpret its magnitude. For example, the Chamfer distance value of 0.002 for the unseen objects is about double the amount of the training objects' average state estimation error; however, in visualization, the reconstructions can reasonably predict the ground truth's curvature. We observed that the majority of errors are due to small undesirable artifacts at the edges of objects which do not significantly impact the general trend of deformations.

\vspace{-5pt}
\subsection{Generalization to Novel Environments \label{appendix - generalization to novel environments}} 
\vspace{-5pt}

\begin{figure}[thpb]
  \centering
    \includegraphics[width=0.8\textwidth]{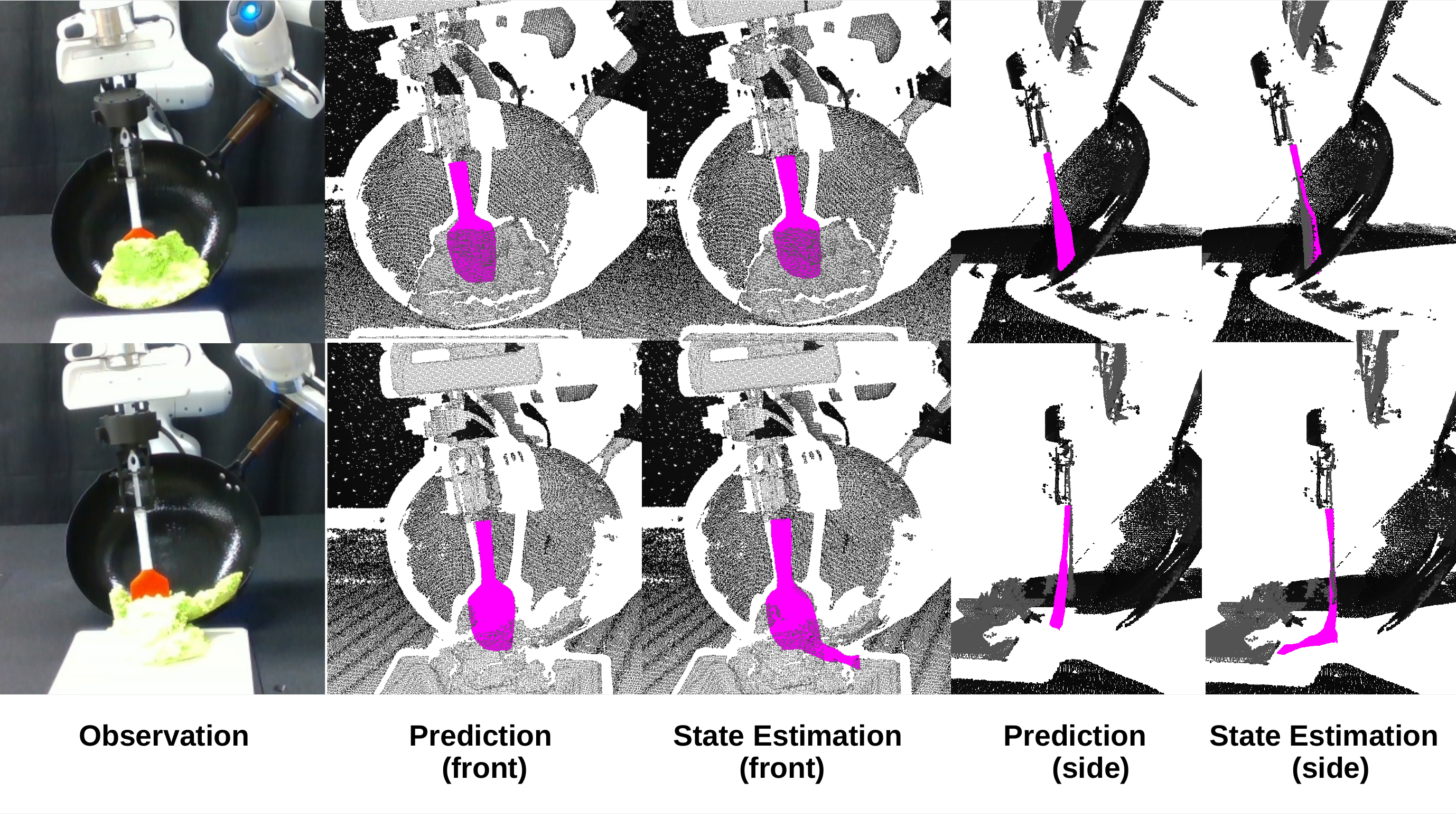}
    \vspace{-10pt}    
    \caption{\small
    \accro can reconstruct high-fidelty dense geometry of deformable objects under occlusions in new environments, such as scraping objects off a wok (successes, top row), but can also at times produce undesirable artifacts (failure modes shown in bottom row).}
    \vspace{-10pt}
  \label{fig:wok-scraping}
\end{figure}

In this section, we show \accro generalization capability to novel environments. Here, the novel environment is a wok scraping task. The wok presents high curvatures producing greater variances in contact formations with respect to the flat table. Here, we add a training dataset with 2 scraping trajectories on the wok and perform state estimation and predictions given a test trajectory with 13 transitions. As in Fig. \ref{fig:wok-scraping}, \accro provides high-fidelity dense geometry of deformable objects in high occlusion cases without a 3D model or configuration of the environment; however, due to the complex environment, it occasionally produces undesired artifacts (2 out of 13 trans.).

\subsection{Ablation study on design choices}
\subsubsection{Explicit Geometry Representation} \label{sec: explicit geometry representation}


\rb{In this section, we compare \accro with explicit models by replacing the geometry representations (deformation and object module) with  GRNet's pointcloud encoder-decoder structure  \cite{xie2020grnet}. Here, we use the same force module as \accro because the baseline does not have a multi-modal sensory encoder. The explicit dynamics model directly takes in a partial pointcloud at time $t$ to generate a dense pointcloud at time $t+1$. The inputs to the dynamics model are pointcloud at the current time step $P_t$, force code $z_t$, \cntemb $c_t$, and action $a_t$ and the output is full pointcloud at the next time step $P_{t+1}$. While $P_t$ is directly input to \cite{xie2020grnet}'s encoder, and other inputs are concatenated to the bottle-neck features.}

\rb{Here, we used the same dataset in Sec.~\ref{sec: experiments - spatula scraping}. 
For the loss function, we replaced $\mathcal{L}^{geo}_{t}$ to Chamfer distance as \cite{xie2020grnet} proposed. The second column of Tab.~\ref{tab: ablation w/ different architectures} shows the result of the explicit dynamics model based on \cite{xie2020grnet}, where the learned explicit model showed better result than \accro for training trajectories. However, the performance dropped significantly given the pointcloud with occlusions from the test trajectories. We speculate that because the explicit dynamics model directly consumes the partial pointcloud to roll out the dynamics, it is more susceptible to visual outliers caused by occlusions. However, \accro is more robust to occlusions due to its implicit nature, using  partial pointcloud implicitely via solving the optimization problem in Sec.~\ref{sec: state estimation}}

\rb{Besides dealing with occlusions, \accro's implicit dynamics model is also advantageous over the explicit counterparts in multi-object interactions. Specifically, the detection of contact and collision at a query point can easily be done by evaluating the sign-distances from each object's signed distance field at a time complexity of $\mathcal{O}(number ~of ~objects)$. On the other hand, explicit geometric representations require special contact/collision detection techniques such as k-d tree \cite{schauer2015collision} or sphere coverings \cite{klein2004point}. These methods have higher time complexities dependent on the length of each point cloud. \cite{weller2013brief, klein2004point,schauer2015collision}. In Sec.\ref{sec: contact location detection}, we demonstrate the accurateness of \accro for detecting contact location.}


\subsubsection{Implicit Geometry Representation with Different Model Architectures}

In this section, we evaluate variations of VIRDO's geometry representation using DeepSDF \cite{park2019deepsdf}, SIREN \cite{sitzmann2020implicit}, and a positional encoder \cite{tancik2020fourier}. In this ablation study we first attempted to replace the \accro's entire geometry representation modules (i.e., both deformation and shape module) with \cite{park2019deepsdf}, \cite{sitzmann2020implicit}, and \cite{tancik2020fourier}, then train each model end to end with the \accro's force and action modules. Under this condition, we set the input as $(\vec{x}, \alpha, z_t)$ and the output as a signed distance function$s \in \mathbb{R}$, where $\vec{x}$ is the query point instead of the entire pointcloud as Sec.~\ref{sec: explicit geometry representation}. However, training end to end without residual layers resulted in failure of convergence for all variations.

We observed that fixing the weights of the  pre-trained object modules and learning only the residuals (deformations) significantly helped the models converge. We chose to replace only the deformation module with the proposed variations and compare performance. The inputs are $(\vec{x}, \alpha, z_t)$ and the output is $\Delta \vec{x}$. We used the same dataset and loss functions as \accro except for training DeepSDF, where we removed $\mathcal{L}_{hyper}$ when training DeepSDF as the architecture does not have a hyper-network.

\textbf{Template-based DeepSDF:} For an apple-to-apple comparison, we set the number of hidden layers = 2 and hidden features = 256 for DeepSDF structures to the same size in \accro's deformation module. We observe that Chamfer distance error becomes  $ 6.54 \sim 6.82$ time greater for the training trajectories and $9.18 \sim  10.72$ times greater for the unseen trajectories when changing the hypernetwork to the DeepSDF architecture. The hypernetwork structure has higher representation power, resulting in more precise geometry reconstructions.

\rb{\textbf{SIREN:} Here, we replaced the geometry representation with SIREN's hypernetwork and trained it with the force and action modules. As suggested in the paper, we implemented SIREN with 5 layers and hidden features of size 256. HyperNetwork (ReLU MLP) was initialized with  Kaiming initialization, and the HypoNetwork was initialized as $w_i \sim \mathcal{U}(-1/l, +1/l)$, where $w_i$ is the network's weight and $l$ is the total number of network parameters. We found that the sine activation function generates bumpy textures resulting in high reconstruction errors for geometry estimation and predictions. }

\rb{\textbf{Positional Encoder:} For this experiment, we add a positional encoder from \cite{tancik2020fourier} right before \accro's deformation modules. Similar to what we observed from SIREN, adding sine or cosine non-linearity to the network can produce undesirable artifacts. We found that the positional encoder generate floating blobs in the free space, resulting in much higher errors than the original \accro.  
 }

\begin{table}
    \centering
        \begin{tabular}{lcccccc}\toprule
        CD ($\times 10^3$) &  \accro  & GRNet & DeepSDF & SIREN & Pos. Enc. \\ \midrule
        Train Est.  & 0.777 (0.215)  &  -- &  5.087 (0.737) & 5.521  (1.015) & 5.569 (2.847)\\
        Train Pred.  &  0.830 (0.330) & 0.759 (0.667) &  5.667 (1.306) &5.457 (0.888) &  5.886 (2.561) \\  \midrule
        Test Est.  & 0.934 (0.210)  & --  &  10.016 (3.253) & 6.349 (1.388) & 8.081 (3.790)\\
        Test Pred. & 1.041 (0.348) &  1.230 (0.826) & 9.552 (2.767) & 6.988 (1.280) & 7.188 (3.603)\\  

        \bottomrule
        \end{tabular}
    \caption{\small \rb{ Results are evaluated in the real-world, showing that \accro outperforms its implicit and explicit counterparts.  
    } } 
    \label{tab: ablation w/ different architectures} 
\vspace{-15pt}
\end{table}


\end{document}